\documentclass[11pt,onecolumn]{IEEEtran}
\pdfoutput=1
\IEEEoverridecommandlockouts

\ifCLASSINFOpdf
\else
\fi
\usepackage{amsfonts}
\usepackage{psfig}
\usepackage{epsf}
\usepackage{bm} 
\usepackage{slashbox}
\usepackage{booktabs}
\usepackage{pifont}
\usepackage{floatrow}
\usepackage{times,amsmath,color,amssymb,graphicx,geometry, epsfig,psfrag,algorithm}
\usepackage{enumerate}
\usepackage{diagbox}
\usepackage{epstopdf}
\usepackage{color}
\usepackage{subfigure}
\usepackage{float}

\usepackage{multirow}

\begin{document}
\title{A Unified Cognitive Learning Framework for Adapting to Dynamic Environment and Tasks}
\author{\IEEEauthorblockN{Qihui Wu, Tianchen Ruan, Fuhui Zhou,  Yang Huang, Fan Xu, Shijin Zhao, Ya Liu,\\ and Xuyang Huang}
\thanks{Q. Wu, T. Ruan, F. Zhou, Y. Huang, F. Xu, S. Zhao, Y. Liu, and X. Huang  are with the College of Electronic and Information Engineering, Nanjing University of Aeronautics and Astronautics, Nanjing, 210000, P. R. China (wuqihui2014@sina.com, rtc0622@sina.com, zhoufuhui@ieee.org, yang.huang.ceie@nuaa.edu.cn, xufan@nuaa.edu.cn, shijin\_zhao@163.com, yaliu\_nuaa@163.com and hxyllxz@163.com).

}
\thanks{The research reported in this article was supported by the Natural Science Foundation of China under Grants 61631020, 61827801 and 61931011.}
}
\maketitle
\begin{abstract}
Many machine learning frameworks have been proposed and used in wireless communications for realizing diverse goals. However, their incapability of adapting to the dynamic wireless environment and tasks and of self-learning limit their extensive applications and achievable performance. Inspired by the great flexibility and adaptation of primate behaviors due to the brain cognitive mechanism, a unified cognitive learning (CL) framework is proposed for the dynamic wireless environment and tasks. The mathematical framework for our proposed CL is established. {\color{blue}Using the public and authoritative dataset, we demonstrate that our proposed CL framework has three advantages, namely, the capability of adapting to the dynamic environment and tasks, the self-learning capability and the capability of ``good money driving out bad money'' by taking modulation recognition as an example.} The proposed CL framework can enrich the current learning frameworks and widen the applications.
\end{abstract}
\IEEEpeerreviewmaketitle
\section{Introduction}
\IEEEPARstart{M}{achine} learning (ML) has received an increasing attention and made great development in wireless communications \cite{C. Jiang}. It enables wireless communication systems to automatically learn and improve performance from experience without being explicitly programmed \cite{Z. Chen}. Note that the traditional ML algorithms significantly depend on a large amount of expert knowledge and require a large amount of high-quality annotated data \cite{Hospedales}. Moreover, a proper set of models and parameters is of great importance for the traditional ML algorithms. In this case, once the training is complete, the model and parameters are determined, and the ML algorithms can only perform the function that is trained. This results in the inadaptability of the new task and practically dynamic wireless environment \cite{Hospedales}. For example, it is difficult to design resource allocation schemes that can adapt to dynamic wireless environment and the changing user task requirement by using the existing machine learning frameworks.

In order to overcome this challenge, the ML algorithms need to learn from the previous experiences and continuously accelerate and improve the learning ability for new tasks. Along this line of thought, inspired by educational psychology, meta-learning has been proposed in machine learning and statistics \cite{Vanschoren}. It is also known as ``learning to learn''. It learns from the previous experiences (also called meta data), which are obtained from the observation of the performance achieved by different machine learning methods performed on a wide range of learning tasks. When the task that exists in the experiences changes, it only needs to fine-tune the parameters. Thus, it can adapt to the  change of task. In \cite{Vanschoren}, a general framework for meta-learning was presented. The most key components of this framework are meta-features and meta-knowledge base. Due to its advantages, meta-learning has been widely used for algorithm selection and hyper-parameter optimization in wireless communications, such as traffic predication \cite{He}, multiple-input and multiple-output (MIMO) detectors \cite{Zhang}, etc.

It is worth noting that when a bran-new task that does not exist in the experiences comes, the performance achieved by using meta-learning is poor and even meta-learning cannot work \cite{Zinkevich}. Moreover, meta-learning requires a large number of task data sets and usually assumes that tasks are independently and identically distributed \cite{ Shalev-Shwartz}. It ignores the situation of non-stationary distribution. If the task data changes dynamically and does not have the same distribution, meta-learning cannot adapt to the change timely \cite{ Shalev-Shwartz}. Furthermore, meta-learning cannot exploit the performance information obtained from the actual tests to improve the learning model. In this case, it is difficult for meta-learning to tackle complex data and complex learning environment, such as dynamic wireless communication environment \cite{Hospedales}. Online learning considers that the training data is sequently and continuously obtained \cite{Zinkevich}. It can quickly adjust the model based on the feedback data, which improves the accuracy of the model. However, the online learning process only focuses on optimizing the current problem. When a new task arrives, the learning speed and accuracy may be decreased since the previous task information is not exploited to obtain an initial parameter of the model \cite{Shalev-Shwartz}.

The authors in \cite{Finn} have combined meta-learning with online learning and proposed online meta-learning, which uses the previous experience to obtain a priori knowledge and can adapt to the current task. However, similar to meta-learning, online meta-learning can only adapt to the change of task that exists in the experience and cannot adapt to the  bran-new task and environment. Moreover, the performance of the online meta-learning is significantly decreased when there exist bad training samples in the meta-knowledge base \cite{Finn}.

To the authors' best knowledge, there are no ML frameworks that not only can quickly adapt to the dynamic environment and new tasks with the help of the increase of new knowledge, such as experience, but is robust to the bad data, e.g. mislabeled data or outliers.  Motivated by the brain cognitive mechanism \cite{ManteContext-dependent2013,SiegelCortical2015} that enables  primate to quickly adapt to dynamic environment and execute complex plans when exposes to new environment and tasks, a unified cognitive learning (CL) framework is proposed in this paper. The mathematical framework for our proposed CL framework is also established. There are three advantages of our proposed CL framework, namely, the ability of adapting to the change of the dynamic environment and tasks, the ability of self-learning and the capability of ``good money driving out bad money''. {\color{blue} We demonstrate those advantages by taking algorithm and hyper-parameters selection as examples with the public and authoritative dataset for modulation recognition.}

The remainder of this paper is organized as follows. Section II presents our proposed cognitive learning. The simulation results are presented in Section III. Finally, Section IV concludes the paper.
\section{Cognitive Learning}

\subsection{Brain Cognitive Mechanism}
\begin{figure}[htb]
\centering
    \subfigure[]
    {
	   \includegraphics[width=0.45\textwidth]{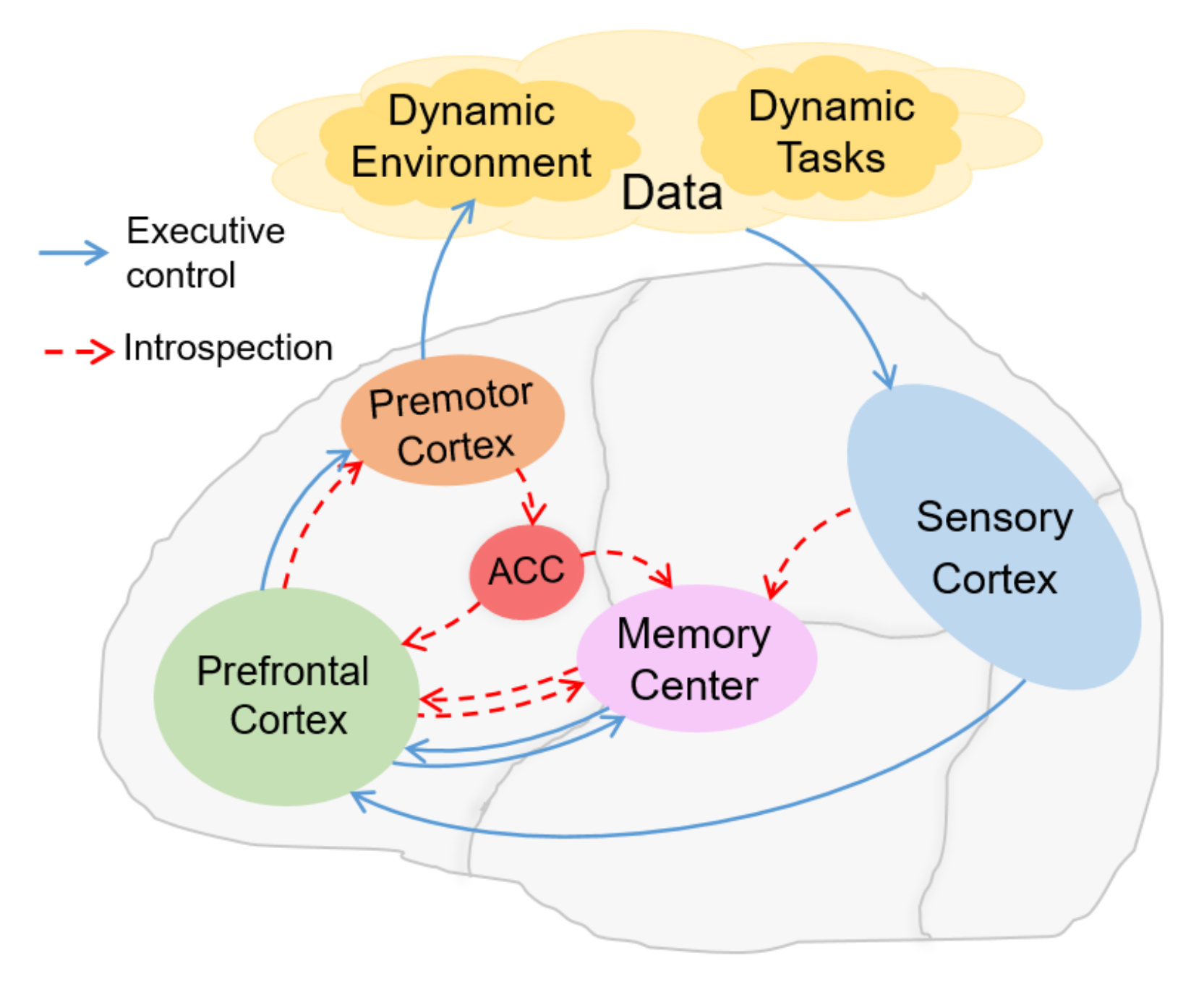}
       \label{Fig.brain.a}
    }
    \subfigure[]
    {
        \includegraphics[width=0.45\textwidth]{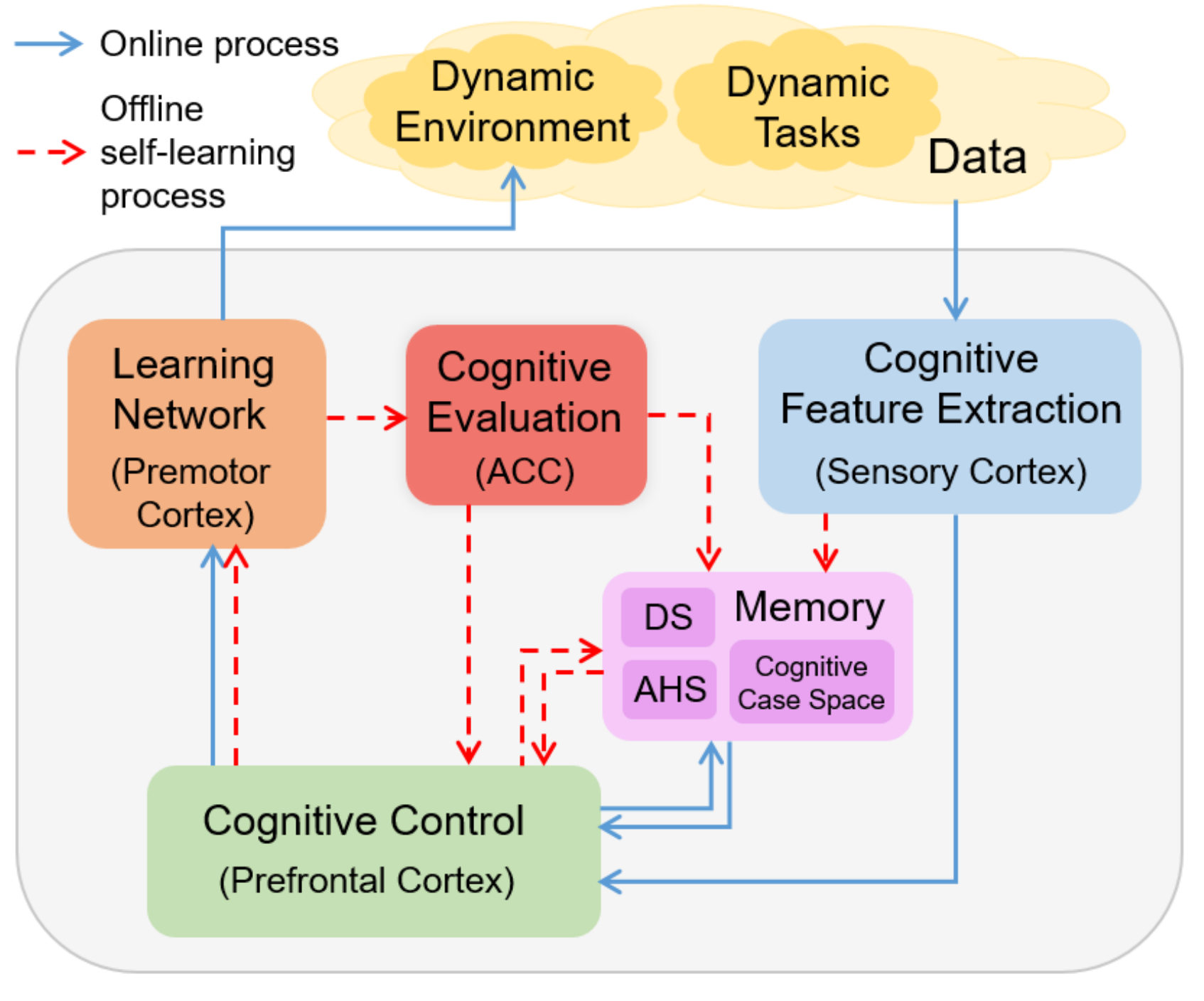}
        \label{Fig.brain.b}
    }

\caption{\textbf{Overview of the brain cognitive mechanism and CL framework.} \textbf{a}, The brain cognitive mechanism. The yellow area represents the outside dynamic environment and dynamic task while the gray brain area represents the brain. The brain contains the sensory cortex, the prefrontal cortex, the premotor cortex, the anterior cingulate cortex (ACC) and the memory center. \textbf{b}, Our proposed cognitive learning framework. It can flexibly select the most suitable algorithm and hyper-parameters according to the dynamic tasks and environment. It has five main modules, namely, the cognitive feature extraction, the cognitive control, the learning network, the cognitive evaluation and the memory module. The memory module has three spaces, namely, a data base (DB), a cognitive case base (CCB), and an algorithm and hyper-parameter base (AHB).} \label{Fig.brain}
\end{figure}

The brain cognitive mechanism is shown in Fig. \textcolor{blue}{\ref{Fig.brain.a}}. It consists of two modes, namely, the executive control process and the introspection process \cite{ManteContext-dependent2013}. The executive control is a complex cognitive process that individuals dynamically and flexibly regulate activities of multiple cognitive subsystems during the goal-directed behavior process. It operates on external tasks and data, and is associated with two broad types of cognitive manipulation. Specifically, the body makes plans to guide behavior. And in case of accidents, the body timely changes gear, which is identified as the rapid adjustment or gating mechanism. This mechanism enables primate to quickly switch behaviors in the dynamic environment \cite{SiegelCortical2015}. However, the executive control is a rapid process and may result in inappropriate motor responses \cite{MansouriConflict-induced2009}. By introspecting the events stored in the internal memory that result in inappropriate motor responses, the introspection process can perform more appropriate motor responses \cite{SiegelCortical2015}. It is a spontaneous mental activity that has nothing to do with the current task or the sensory environment. The details of the executive control and the introspection process are presented as follows.

As the blue solid line shown in Fig. \textcolor{blue}{\ref{Fig.brain.a}}, the executive control process has four steps. Firstly, the sensory cortex receives data from the dynamic environment and tasks and performs feature extraction from them. Secondly, the prefrontal cortex (PFC) integrates features from the sensory cortex and retrieves the related experience knowledge from the memory center. Thirdly, based on the feature information and the related experience knowledge, the PFC exerts cognitive control in order to obtain the stimulus-response association information. Finally, the premotor cortex obtains the specific motor plans based on the stimulus-response association information. Those motor plans are responded to the dynamic environment.

As the red dot line shown in Fig. \textcolor{blue}{\ref{Fig.brain.a}}, the introspection process has five steps. Firstly, features of the dynamic environment and tasks are stored in the memory center. Secondly, the PFC integrates features from the memory center and retrieves the related experience knowledge from the memory center. The third and fourth steps of the introspection process are similar to those of the executive control process. The only difference is that the motor plans are exported to the anterior cingulate cortex (ACC). Finally, the ACC monitors the conflict response in the premotor cortex and feeds back the conflict information to the PFC. The stimulus-response association information is stored in the memory center. In this case, PFC can regulate cognitive control and attention resources, and a better decision can be made. Thus, introspection can correct inappropriate responses and make better decisions.

\subsection{Cognitive Learning Framework}
Motivated by the brain cognitive mechanism, as shown in Fig. \textcolor{blue}{\ref{Fig.brain.b}}, a CL framework is proposed to achieve the adaptive capability for the dynamic environment and dynamic tasks and the self-learning capability for making better decisions. Note that in Fig. \textcolor{blue}{\ref{Fig.brain.b}}, the modules used in the same color are correspondent. Similar to the brain cognitive mechanism, our proposed CL framework also consists of two processes, namely, the online process and offline self-learning process that are corresponding to the executive control and introspection process of the brain cognitive mechanism, respectively.

As the blue solid line shown in Fig. \textcolor{blue}{\ref{Fig.brain.b}}, similar to the executive control process of the brain cognitive mechanism, the online process of our proposed CL framework also has four steps. Firstly, the cognitive feature extraction module extracts features from the dynamic environment and tasks. Secondly, the cognitive control module addresses those features and establishes the matching relationship between the obtained features and the selection of the appropriate algorithm type and hyper-parameters in order to obtain the appropriate algorithm type and hyper-parameters. Thirdly, based on the selected algorithm type and hyper-parameters, the cognitive control module acquires the specific algorithm and hyper-parameter values from the memory module, and then reconstructs the selected algorithm. Finally, the learning network module performs the selected algorithm based on the data and obtains the learning results. The results are responded to the dynamic environment.

The online process cannot guarantee to obtain the most appropriate algorithm type and hyper-parameters. In contrast, as the red dotted line shown in  Fig. \textcolor{blue}{\ref{Fig.brain.b}}, the offline self-learning process continually updates the matching relationship between features of the dynamic environment and tasks and more appropriate selection of the algorithm type and hyper-parameters in order to obtain the most appropriate algorithm type and hyper-parameter values. Specifically, the cognitive evaluation module in the offline self-learning process compares the current learning result with the previous learning results. Then, the cognitive control module adjusts the selection of the algorithm type and hyper-parameters based on the performance evaluation result. If the performance of the current learning result is the best, that is, the offline self-learning process obtains the most appropriate algorithm type and hyper-parameter values, the offline self-learning process is complete. Otherwise, this process is continually performed. Thus, the offline self-learning process can improve the algorithm and hyper-parameter selection performance.

The pivotal modules of our proposed CL framework are the cognitive feature extraction, cognitive control, cognitive evaluation and cognitive case space. The cognitive feature extraction not only can obtain the feature of the dynamic environment and tasks, but can reflect the dynamic change of the environment and tasks, which are helpful for the cognitive control module to quickly select the appropriate candidate algorithm types and hyper-parameters when the environment and tasks change. The cognitive control establishes the matching relationship between the obtained features of the dynamic environment and tasks and the selection of the appropriate algorithm type and hyper-parameters, which enables our proposed CL framework to adapt to the change of the environment and task. Moreover, in the offline self-learning process, the  matching relationship can be continuously updated and thus the knowledge can be cumulated, which are beneficial for selecting the most appropriate algorithm type and hyper-parameters. The cognitive evaluation module evaluates the performance of the selected algorithm type and hyper-parameters, which enables the cognitive case space to accumulate better knowledge.  For the cognitive case space, the knowledge about the relationship between features of the dynamic environment and tasks and the selected algorithm and hyper-parameter can be cumulated, which decreases the impact of the bad knowledge about the inappropriate matching relationship.
\subsection{Mathematical Framework for Cognitive Learning}
\begin{figure}[htb]
\centering
\includegraphics[width=1\textwidth]{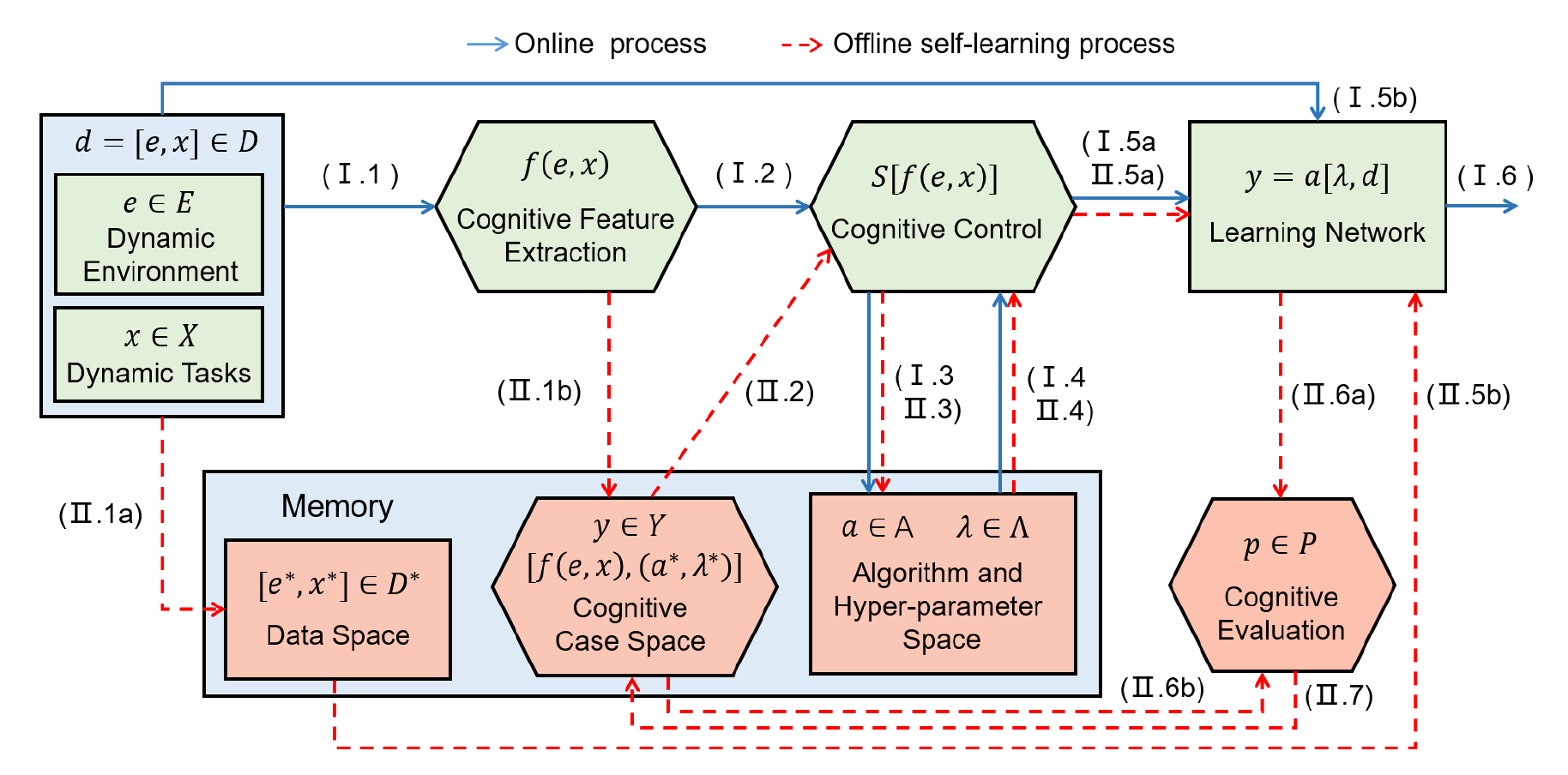}
\caption{Mathematical framework for our proposed cognitive learning.}
\label{Fig.mathframe}
\end{figure}

{\color{blue}Based on our proposed CL framework, a mathematical framework for CL is established, as shown in Fig. \textcolor{blue}{\ref{Fig.mathframe}}. This framework is obtained from the mathematical framework presented in \cite{lecun1998gradient}. It consists of online and offline self-learning process, which are respectively represented by the solid line and dotted line arrow. The denotations of operations involved in our proposed CL mathematical framework is shown in Table 1.}

\renewcommand{\thetable}{\arabic{table}}
\renewcommand{\tablename}{Table}
\begin{table}[!htbp]

\centering
 \caption{\label{tab:test}The denotations of the operations.}
 \resizebox{\textwidth}{!}{
\begin{tabular}{|l|l|}
  \hline
  Online process  & Offline self-learning process \\ \hline
  \uppercase\expandafter{\romannumeral1}.1 Input data of dynamic environment and tasks& \uppercase\expandafter{\romannumeral2}.1a Store the data of dynamic environment and tasks\\ \hline
  \uppercase\expandafter{\romannumeral1}.2 Transmit features of dynamic environment and tasks& \uppercase\expandafter{\romannumeral2}.1b Store features of dynamic environment and tasks\\ \hline
  \uppercase\expandafter{\romannumeral1}.3 Operate on algorithm and hyper-parameter base& \uppercase\expandafter{\romannumeral2}.2 Transmit features of dynamic environment and tasks\\ \hline
  \uppercase\expandafter{\romannumeral1}.4 Extract algorithm and hyper-parameters& \uppercase\expandafter{\romannumeral2}.3 Operate on algorithm and hyper-parameter base\\ \hline
  \uppercase\expandafter{\romannumeral1}.5a Transmit the selected algorithm and hyper-parameters& \uppercase\expandafter{\romannumeral2}.4 Extract algorithm and hyper-parameters \\ \hline
  \uppercase\expandafter{\romannumeral1}.5b Input data of dynamic environment and tasks& \uppercase\expandafter{\romannumeral2}.5a Transmit the selected algorithm and hyper-parameters\\ \hline
  \uppercase\expandafter{\romannumeral1}.6 Output the learning result& \uppercase\expandafter{\romannumeral2}.5b Input data of dynamic environment and tasks\\
  \hline
  & \uppercase\expandafter{\romannumeral2}.6a Transmit current learning result\\
  \hline
  & \uppercase\expandafter{\romannumeral2}.6a Transmit previous learning results\\
  \hline
  & \uppercase\expandafter{\romannumeral2}.7 Transmit better learning and algorithm hyper-parameters\\
  \hline
\end{tabular}
}
\end{table}

As shown in Fig. \textcolor{blue}{\ref{Fig.mathframe}}, the input of the mathematical framework is the data (denoted by $d$) related to the dynamic environment and dynamic tasks, which are denoted by $e$ and $x$ respectively. $D$, $E$ and $X$ denote the set of data, dynamic environment and dynamic tasks, respectively. Note that ``dynamic'' means that the environment and tasks are dynamically changed, which may be the same as or different from the existing environment and tasks. Those data of dynamic environment and dynamic tasks (denoted by $e^\ast$ and $x^\ast$) are also stored in the data space of the memory module for the future utilization, where $\ast$ denotes the historical data instead of the real-time data of the environment and tasks. The memory module also has cognitive case space and algorithm and hyper-parameter space. The cognitive case space consists of the learning result set denoted by $Y$ and the cognitive space denoted by $\left[f\left(e, x\right),({a^*}, {\lambda^*})\right]$ where $f\left(e, x\right)$ denotes the feature of the dynamic environment and tasks and $({a^*}, {\lambda^*})$ denotes the selected algorithm type and hyper-parameters. The algorithm and hyper-parameter space consists of the available algorithm type set denoted by $A$ and the hyper-parameter set denoted by $\Lambda$.

Besides the above-mentioned two modules, the mathematical framework have four other modules. Firstly, the cognitive feature extraction module extracts features of the dynamic environment and dynamic tasks denoted by $f\left(e, x\right)$, which are input into the cognitive control module and stored in the cognitive case space. Secondly, the cognitive control module establishes the matching relationship between features of the dynamic environment and dynamic tasks and the selection of the algorithm type and hyper-parameters, denoted by $S\left[f\left(e, x\right)\right]$, which can be updated in the offline self-learning process in order to obtain the selection of the most appropriate algorithm type and hyper-parameters. Thirdly, the learning network module performs the algorithm with the input data and obtains the learning result. Finally, the cognitive evaluation module evaluates the performance of the current learning result based on the previous results, and then feeds the evaluation value $p$ back to the cognitive control module to regulate the selection of the algorithm type and hyper-parameters.
\subsection{Advantages of Cognitive Learning}
Compared with the existing learning frameworks, e.g., meta-learning, our proposed CL has three advantages. The details are presented as follows.

{\color{blue}Firstly, CL can exploit features of the dynamic environment and tasks and the learning results stored in the cognitive case space to improve the learning performance. The reason is that the cognitive control module can adjust and update the matching relationship between features of the dynamic environment and tasks and the selection of the appropriate algorithm type and hyper-parameters based on the cognitive evaluation result. In this case, a more appropriate algorithm type and hyper-parameters can be selected. Thus, our proposed CL framework has the ability of self-learning.}

Secondly, since the learning result is influenced by the environment and the task, when the environment or task changes, the learning results can be very different. As shown in Fig. \textcolor{blue}{\ref{Fig.brain.b}}, the cognitive feature extraction module extracts new features of the dynamic environment and task when the environment or task changes. Based on the matching relationship between features of the dynamic environment and task and  the selection of the algorithm type and hyper-parameters, the cognitive control module can change the selection of the algorithm type and hyper-parameters based on the new features. Thus, our proposed CL can adapt to the dynamic environment and dynamic tasks.

Finally, CL has the ability of ``good money driving out bad money''. In the cognitive case space, training samples may have mislabels, that is, for certain task or environment, the algorithm type or the hyper-parameters that are not the most appropriate are labeled as the most appropriate one, which decreases the performance of the learning results. In our proposed CL framework, the cognitive evaluation module can compare the current learning result with the previous results. In this case, the good training samples can be stored in the cognitive case space and the bad training samples can be reduced, which improves the performance of the learning result.
\section{Simulation Results}
{\color{blue}Simulation results are given to compare our proposed CL framework with the meta-learning (MtL) framework  in terms of three key capabilities, namely, the capability of self-learning, the capability of adapting to dynamic environments and tasks, and the capability of ``good money drives out bad money''.  Those results are obtained by modulation recognition conducted on the public and authoritative datasets and the selection of the  modulation recognition algorithm and hyper-parameters (HP) is taken as examples. The RadioML 2016.10A dataset is used \cite{P. Qi} \cite{S. Huang}, which contain 11 different modulations (8 for digital modulations, 3 for analog modulations) with different signal to noise ratios (SNRs range from -20dB to 18dB with step of 2dB). There are 1000 samples for each group of modulation and SNR. The length of each sample is 128  and is represented by its in-phase and quadrature components.

For our proposed CL framework, the neural network is exploited as the cognitive control module, which establishes the matching relationship between the input features and the selection of the appropriate algorithm type or hyper-parameters. The inputs of the neural network are features of the problem while the output is the selected algorithm or hyper-parameters. In order to obtain the optimal modulation recognition algorithm and hyper-parameters for the modulation recognition problem, features of the problem consist of features of the RF signal dataset (where the modulation classification task comes from) and the performance requirements in terms of modulation classification. The performance requirements consist of the accuracy of modulation recognition and the completion time for modulation recognition. In the simulations, the selection of an appropriate algorithm or hyper-parameters means that the selected algorithm or hyper-parameters can satisfy the performance requirements while the selection of the most appropriate algorithm or hyper-parameters means that the selected algorithm or hyper-parameters obtain the optimal performance for the given RF signal dataset and performance requirements. The modulation recognition  environment change means that SNR, the number of training samples and the number of modulation types in the dataset used for modulation recognition are changed. In the simulation, five sub-datasets from RadioML 2016.10A dataset are generated, namely, dataset1, dataset2, dataset3, dataset4, and dataset5. There are 1000 training samples in dataset1-4 while 500 training samples exist in dataset5; dataset1-3 and dataset5 have 11 modulation types while dataset4 has AM-DSB, BPSK, CPFSK, GFSK, PAM4 and QPSK. The SNRs of dataset1, dataset4 and dataset5 are 18 dB while SNR of dataset2 is -16 dB and that of dataset3 is 2 dB.
\subsection{Capability of Self-Learning}
We perform extensive tests to demonstrate the advantage of our proposed CL framework in terms of the capability of self-learning compared with the MtL framework. Fig. \textcolor{blue}{\ref{Fig.autonomous.a}} and Fig. \textcolor{blue}{\ref{Fig.autonomous.b}} respectively show the performance of the algorithm selection accuracy and modulation recognition accuracy versus the index of the test while Fig. \textcolor{blue}{\ref{Fig.autonomous.c}} and Fig. \textcolor{blue}{\ref{Fig.autonomous.d}} respectively show the performance of the hyper-parameter selection accuracy and modulation recognition accuracy versus the index of the test. The algorithm  or hyper-parameter  selection  accuracy is evaluated by checking whether the selected algorithm or hyper-parameters are the labeled ones or not while the modulation recognition accuracy is evaluated by performing modulation recognition with the selected algorithm or hyper-parameters and checking whether the classification results are correct or not. If the modulation recognition task cannot be completed in the given time (i.e. the running time requirement), the modulation recognition accuracy is set as zero.

\begin{figure}[H]
    \centering
    \subfigure[]
    {
	    \includegraphics[width=0.40\textwidth]{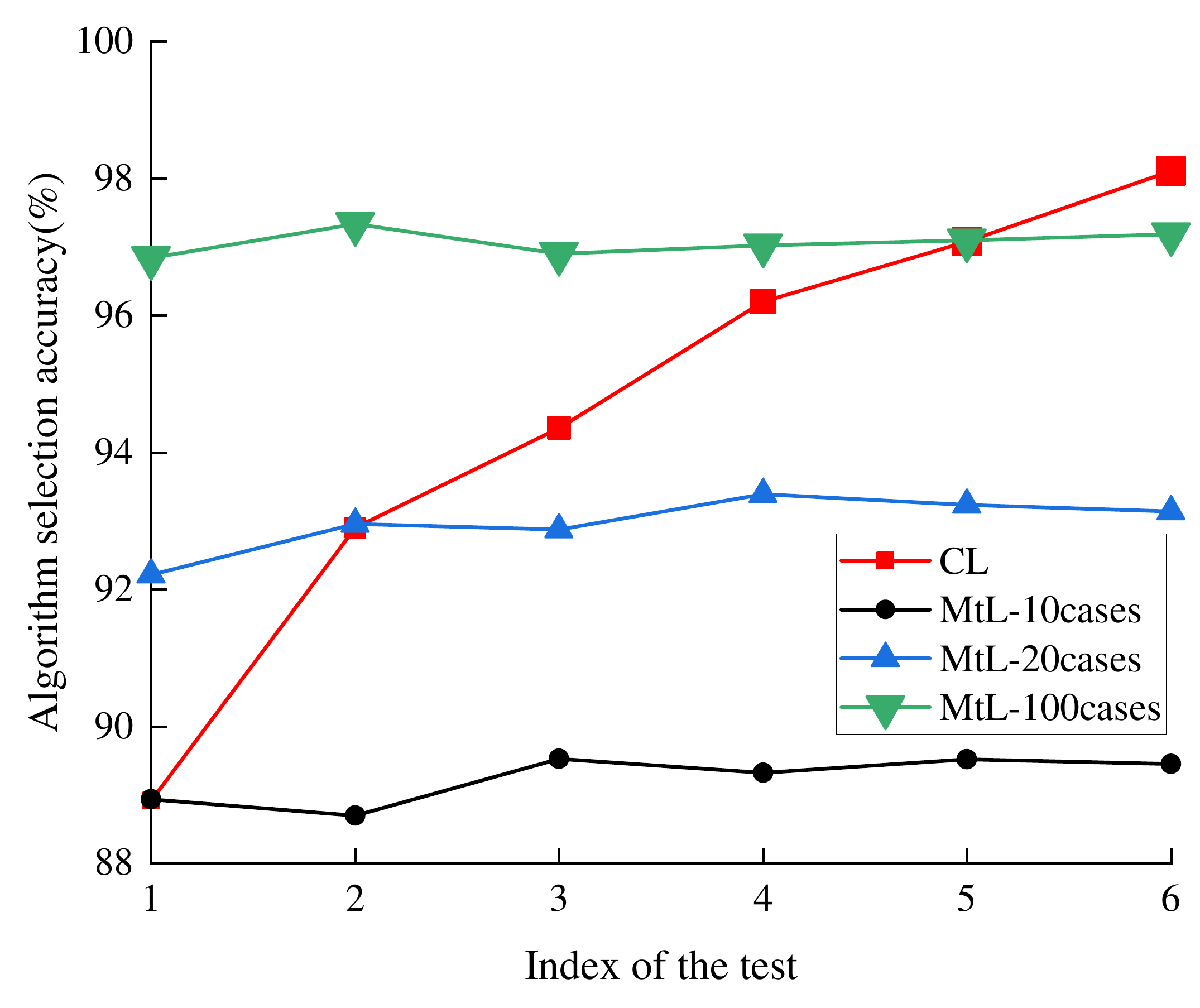}
        \label{Fig.autonomous.a}
    }
    \quad
    \subfigure[]
    {
	   \includegraphics[width=0.40\textwidth]{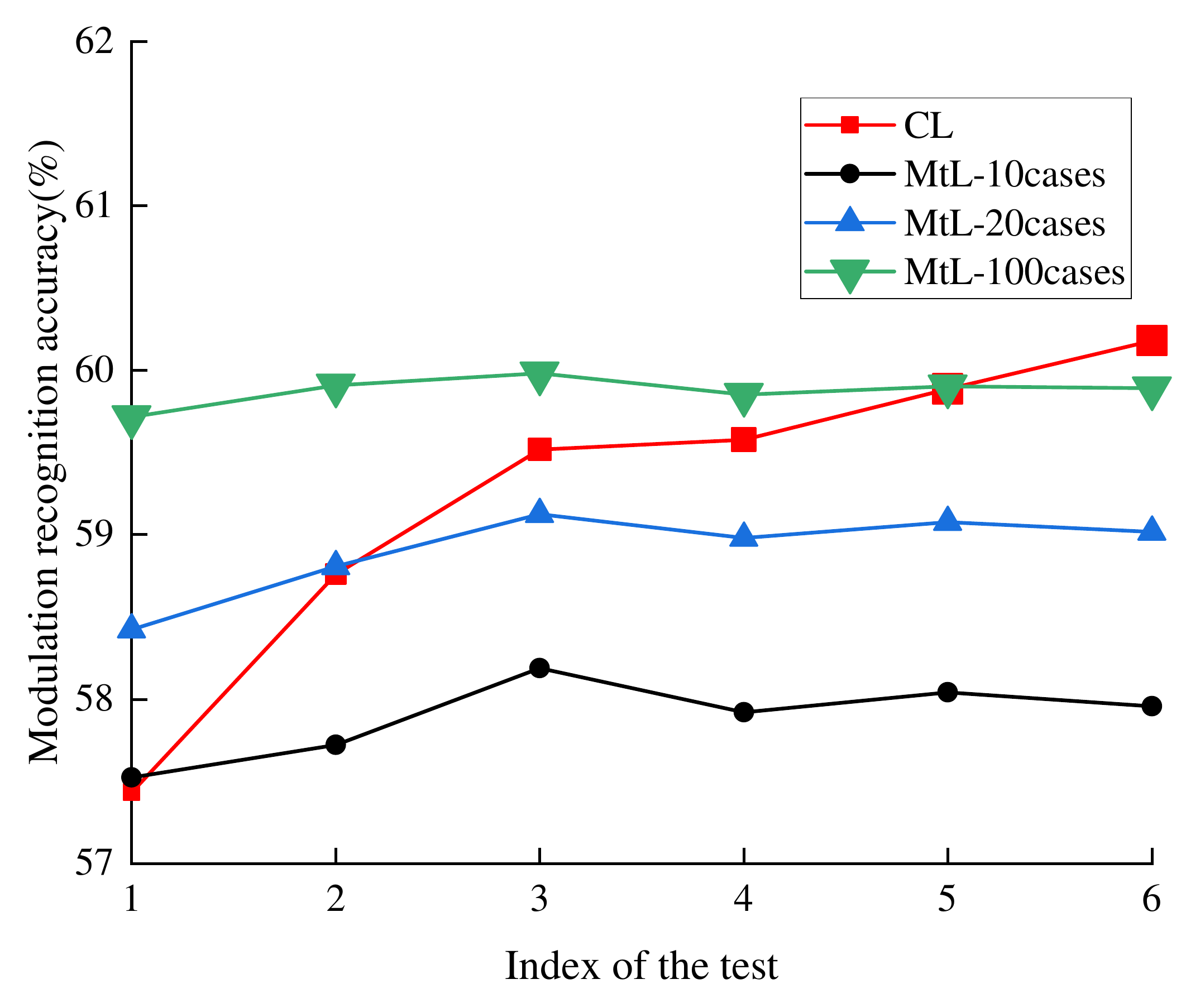}
        \label{Fig.autonomous.b}
    }\vskip 4pt

    \subfigure[]
    {
	   \includegraphics[width=0.40\textwidth]{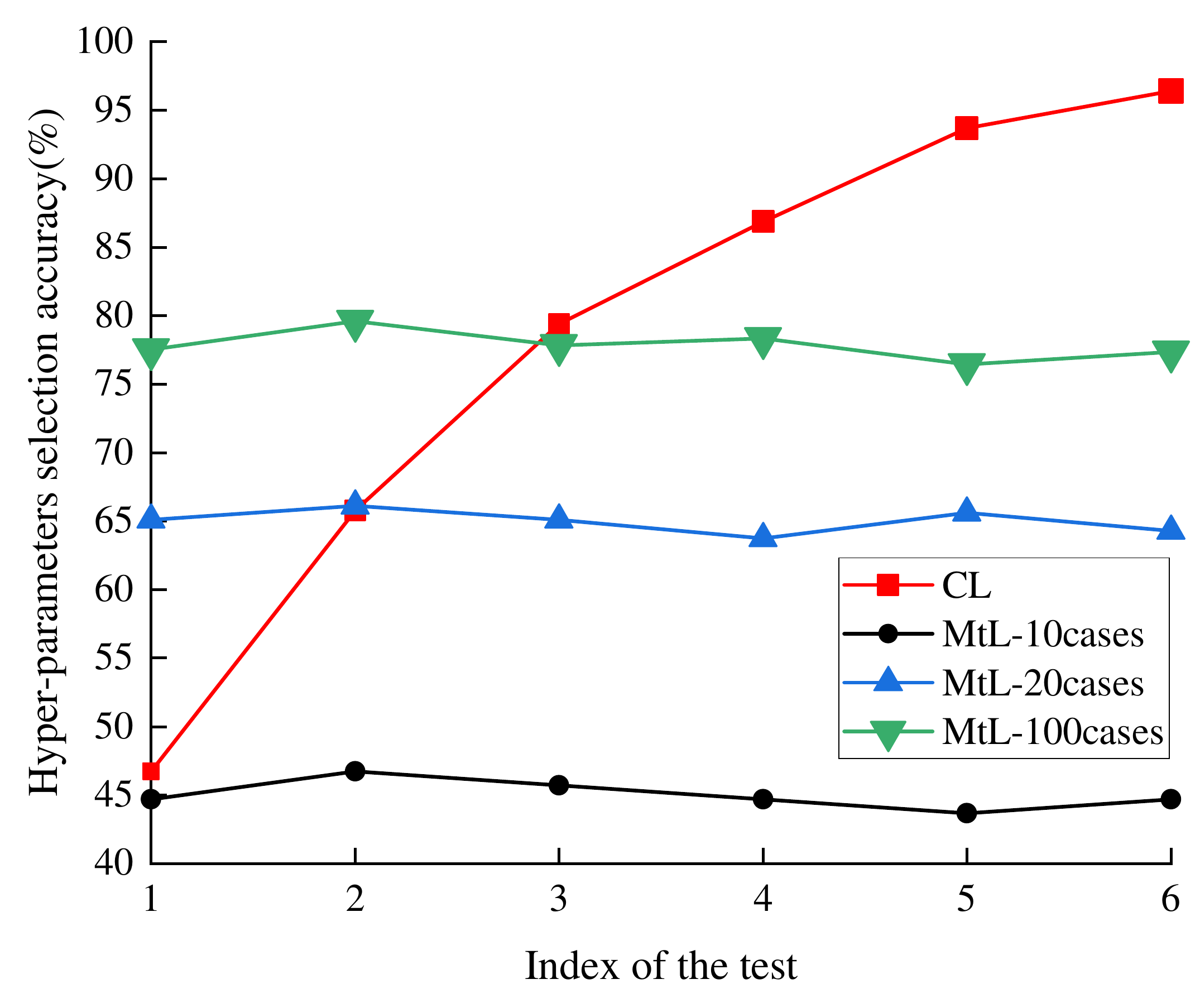}
        \label{Fig.autonomous.c}
    }
    \quad
    \subfigure[]
    {
	   \includegraphics[width=0.40\textwidth]{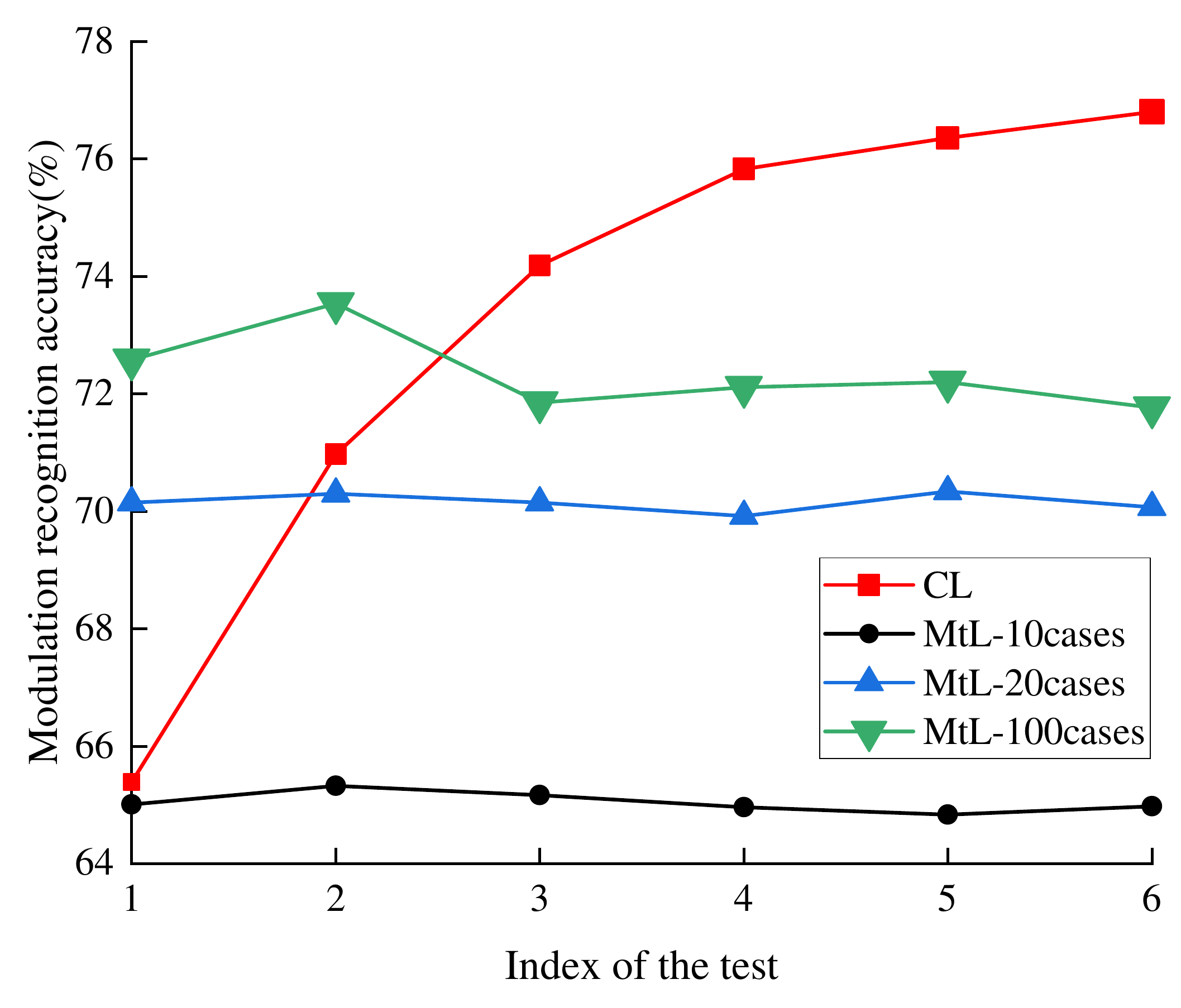}
        \label{Fig.autonomous.d}
    }\\
    \caption{{\textbf{CL versus MtL in terms of the capability of self-learning.} \textbf{a}, Algorithm selection accuracy. \textbf{c}, Hyper-parameter selection accuracy. \textbf{b} and \textbf{d},  Modulation recognition accuracy achieved by the selected algorithm and hyper-parameters, respectively. The increase of the symbols size represents the increase of the number of cognitive cases.}
    \label{Fig.autonomous}}
\end{figure}

Since our proposed CL framework has cognitive case space that can store the test cases, the size of the cognitive case space can be enlarged as the test continuously be conducted. In simulations, before the first test, for each RF signal dataset, 10 cognitive cases are used to train the neural network, and in the first test the selection performance is evaluated by using additional 10 testing cases. Those testing cases are added into the cognitive case space. Thus, in each dataset, there are 20 cognitive cases, which are exploited to train the neural network. In the second test, 10 testing cases are used to evaluate the selection performance. Similarly, those testing cases are then added into the cognitive case space. In Fig. \textcolor{blue}{\ref{Fig.autonomous.a}}, the number of testing cases is  30, 40, 200 and 200 for the 3rd, 4th, 5th, and 6th test, respectively. On the contrary, due to the lack of cognitive case space, the space of meta examples of the  MtL-based  selection remains constant. Therefore, the  MtL-based selection is pre-trained with given meta examples before testing. In Fig. \textcolor{blue}{\ref{Fig.autonomous}a} and Fig. \textcolor{blue}{\ref{Fig.autonomous}b}, the number of the given meta examples is 10, 20 and 100 while in Fig. \textcolor{blue}{\ref{Fig.autonomous}c} and Fig. \textcolor{blue}{\ref{Fig.autonomous}d}, the number of the given meta examples is 10, 20 and 100.

It is seen from Fig. \textcolor{blue}{\ref{Fig.autonomous}} that the algorithm or hyper-parameter selection accuracy and the  modulation  classification accuracy increase with the index of test achieved by using our proposed CL framework while those performances obtained with the MtL framework are almost the same for different indexes when the number of meta examples is constant. This is due to the fact that in contrast to the MtL framework, the CL framework can enlarge the cognitive case space by absorbing testing cases, which are used as training samples to train the neural network. Thus, for the same dataset, the test error of the neural network for the algorithm or hyper-parameter selection can be reduced with the increase of the cognitive cases. However, for the MtL framework, the number of the meta examples is a constant and cannot be increased during the algorithm or hyper-parameter selection process. This result demonstrates that our proposed framework has the capability of self-learning.
\subsection{Capability of Adapting to Dynamic Environment and Tasks}
To demonstrate that our proposed CL framework has the capability of adapting to the dynamic environment and tasks, performance comparison between our proposed CL framework and the MtL framework is provided. The dynamic environment means the change of dataset while the dynamic tasks indicate the change of the performance requirements. In simulations, the change of environment results in the change of features of the RF signal dataset and the changes of the performance requirements are the completion time and the algorithm or hyper-parameter selection accuracy.

Fig. \textcolor{blue}{\ref{Fig.dynamic.a}} and Fig. \textcolor{blue}{\ref{Fig.dynamic.b}}  show the algorithm and hyper-parameter selection accuracy when the environment or tasks change. Those results are achieved by using our CL framework and the MtL framework.  The following results can be both seen from Fig. \textcolor{blue}{\ref{Fig.dynamic.a}} and Fig. \textcolor{blue}{\ref{Fig.dynamic.b}}. Before the first test, the neural network for the algorithm or hyper-parameter selection is pre-trained on dataset1 to satisfy the ``completion time in priority'' requirement.  Based on the pre-trained model for the ``completion time in priority'' requirement, the 1st and the 2nd tests are carried out to evaluate the accuracy of the algorithm or hyper-parameter selection on dataset1. Both frameworks work well. When the requirement changes in the 3rd test, the performances of those two frameworks are significantly decreased, irrespective of the algorithm or hyper-parameters selection. It is also seen that the algorithm selection accuracy achieved by using our proposed CL framework is quickly increased from the 4th to the 7th test and the hyper-parameter selection accuracy achieved by using our proposed CL framework is quickly increased from the 4th to the 9th test and eventually reaches a high level around 90.8\%. However,  the algorithm and hyper-parameter selection accuracy achieved with the  MtL framework remains a low level. Dataset change occurs on the 11th test both for the algorithm selection and the hyper-parameter selection. It is seen that the performance achieved by using our proposed CL framework or the MtL framework is significantly decreased. As the test continues, the performance achieved by using our proposed CL framework is quickly increased and reaches a new high level  while the MtL method stays at original level.
\begin{figure}[H]
    \centering
    \subfigure[]
    {
        \includegraphics[width=0.75\textwidth]{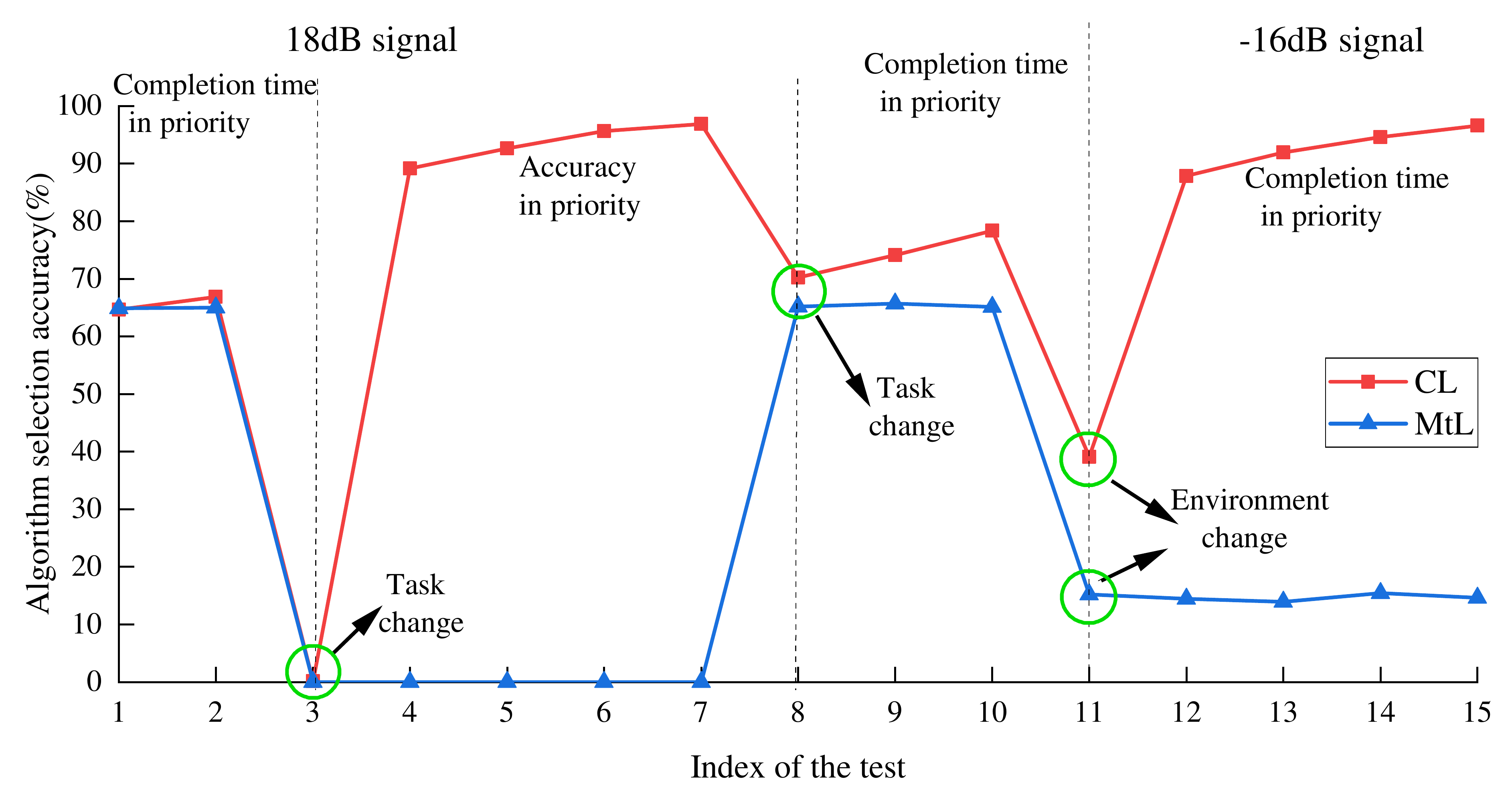}
        \label{Fig.dynamic.a}
    }\vskip 1pt

    \subfigure[]
    {
        \includegraphics[width=0.75\textwidth]{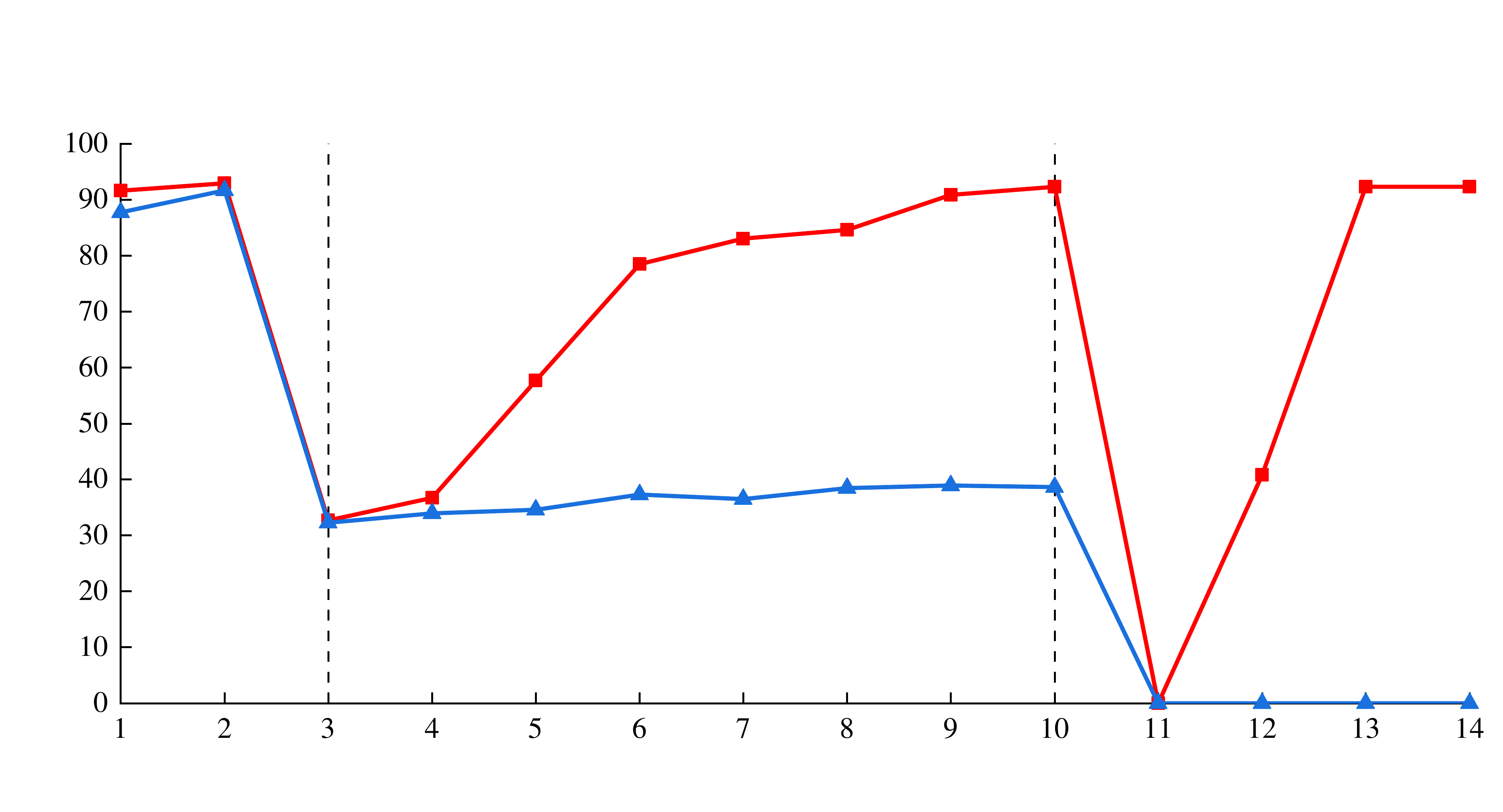}
        \label{Fig.dynamic.b}
    }

    \caption{{\textbf{Cognitive learning versus meta learning in terms of the capability of adapting to the dynamic environment and tasks.} The cycle highlights the changes of tasks and environment. \textbf{a}, Algorithm selection accuracy. \textbf{b}, Hyper-parameter selection accuracy.}
    \label{Fig.dynamic}}
\end{figure}

The above results are due to the fact that for the algorithm or hyper-parameter selection based on our proposed CL framework, the new test cases with respect to a new dataset or a new performance requirement can be added to the cognitive case space. Thus, by training with these new cognitive cases, the new matching relationship between new features of the new environment and new tasks and the selection of the algorithm and hyper-parameters can be added to the neural network for the algorithm or hyper-parameter selection, thus contributing to a better performance for the algorithm or hyper-parameter selection under changes of requirement or dataset. Those results demonstrate that our proposed CL framework has the capability of adapting to the dynamic environment and tasks. On the contrary, since the MtL framework does not have this capability, the performance achieved with the MtL framework is poor when the environment and tasks change.

\subsection{Capability of ``Good  Money Driving Out Bad Money''}

We also performs extensive tests to demonstrate that our proposed CL framework has the capability of  ``good money driving out bad money''.  The bad money refers to the mislabeled cases from the cognitive case space. The degree of bad money is represented by the ratio of the mislabeled cases. In simulations, for the algorithm or hyper-parameter selection, training cases for the algorithm or hyper-parameter selection may be mislabeled, and thus, for given dataset and performance requirements, the output algorithm or hyper-parameters from the neural network for the algorithm or hyper-parameter selection are inappropriate.

We consider two cases in the simulation. One is that the ratio of the mislabeled cases is 10\% and the other is that  the ratio of the mislabeled cases is 30\%. Fig. \textcolor{blue}{\ref{Fig.badsamples.a}} and Fig. \textcolor{blue}{\ref{Fig.badsamples.c}} show the algorithm and hyper-parameter selection accuracy versus the index of the test achieved with our proposed CL framework and those obtained with the MtL framework under different ratios of the mislabeled cases while Fig. \textcolor{blue}{\ref{Fig.badsamples.b}} and Fig. \textcolor{blue}{\ref{Fig.badsamples.d}} show the corresponding modulation recognition accuracy. It is seen that the performance achieved based on our proposed framework is better than that obtained with the MtL framework, irrespective of the algorithm or hyper-parameter selection accuracy, or the modulation recognition accuracy.

Moreover, the performance achieved based on our proposed framework increases with the index of the test while the performance achieved based on the MtL framework fluctuates in a range. Those results can be explained by two aspects. On one hand, in our proposed CL framework, the cognitive evaluation module can compare the current learning result with the pervious results and feeds back the evaluation result to the cognitive control module.

\begin{figure}[H]
    \centering
    \subfigure[]
    {
	   \includegraphics[width=0.45\textwidth]{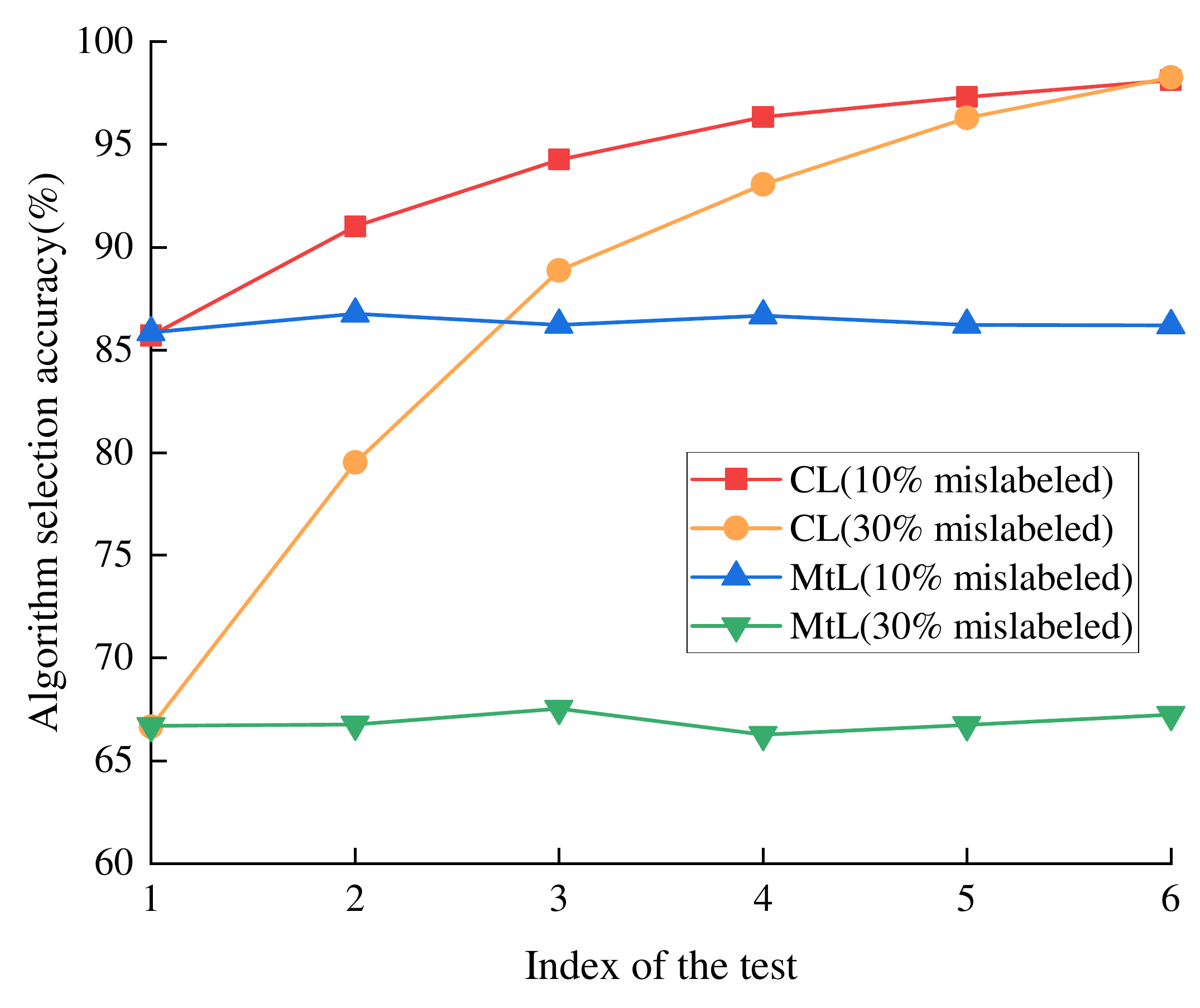}
        \label{Fig.badsamples.a}
    }
    \quad
    \subfigure[]
    {
        \includegraphics[width=0.45\textwidth]{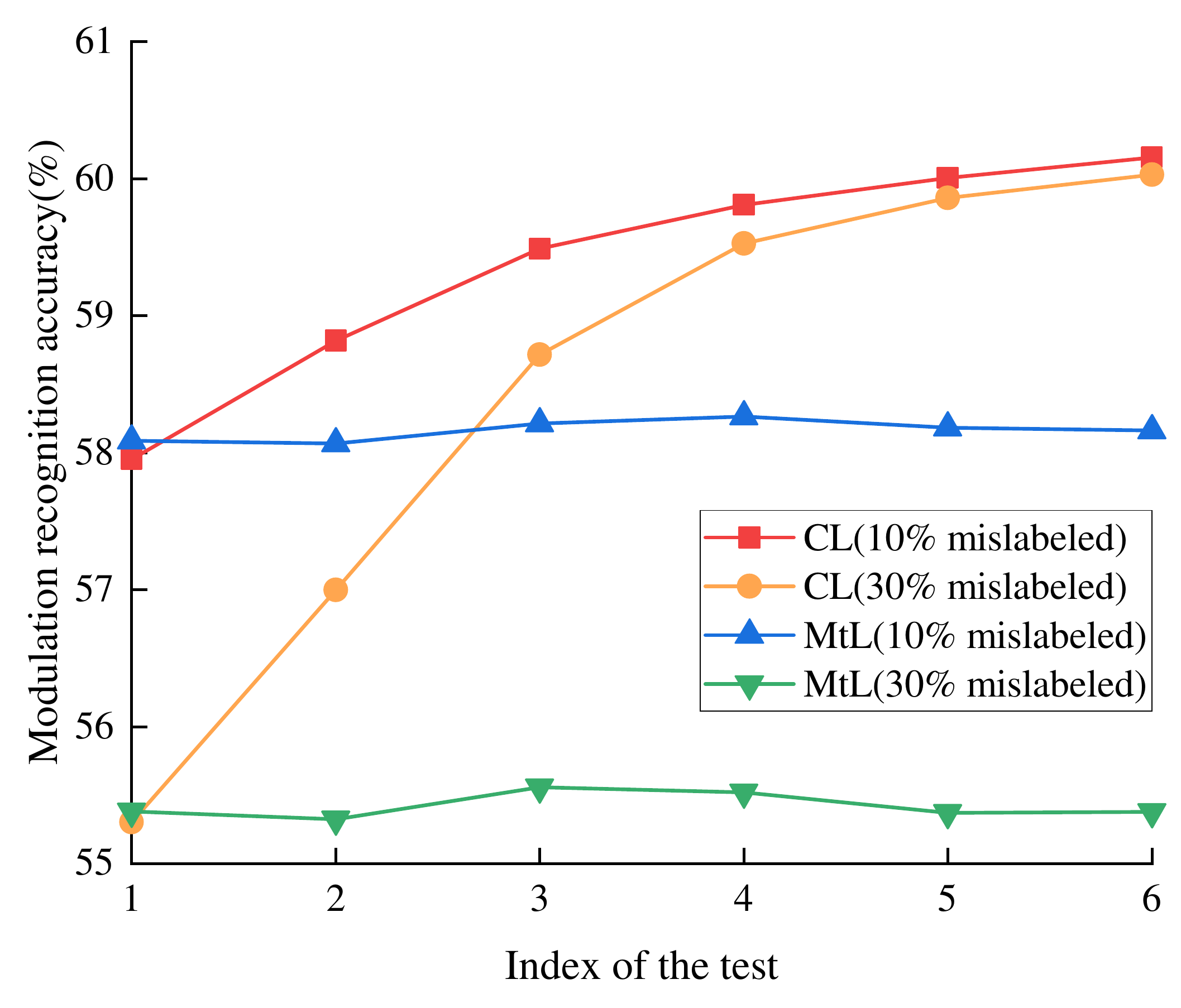}
        \label{Fig.badsamples.b}
    }\vskip 4pt

    \subfigure[]
	{
		\includegraphics[width=0.45\textwidth]{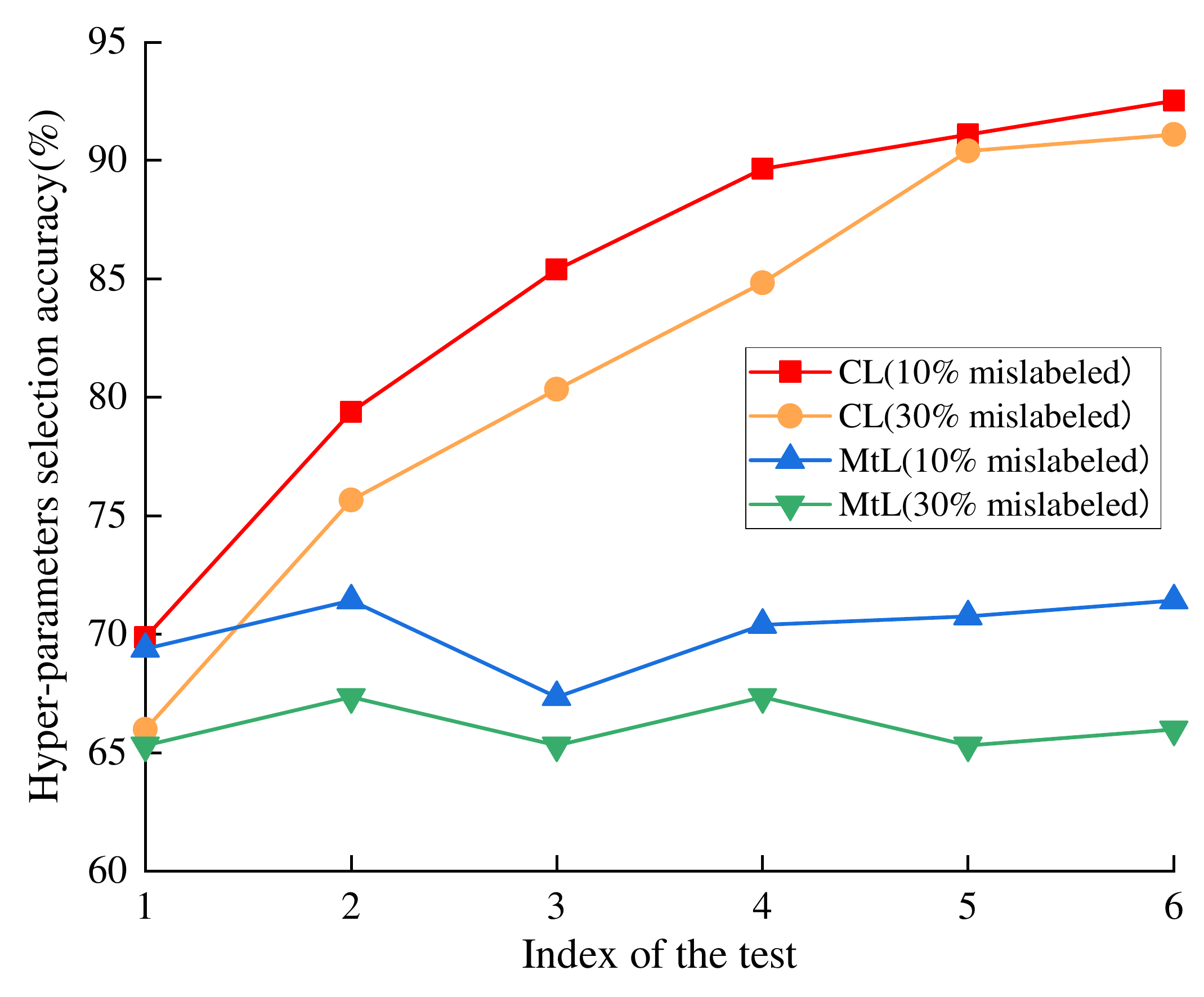}
        \label{Fig.badsamples.c}
	}
    \quad
	\subfigure[]
	{
		\includegraphics[width=0.45\textwidth]{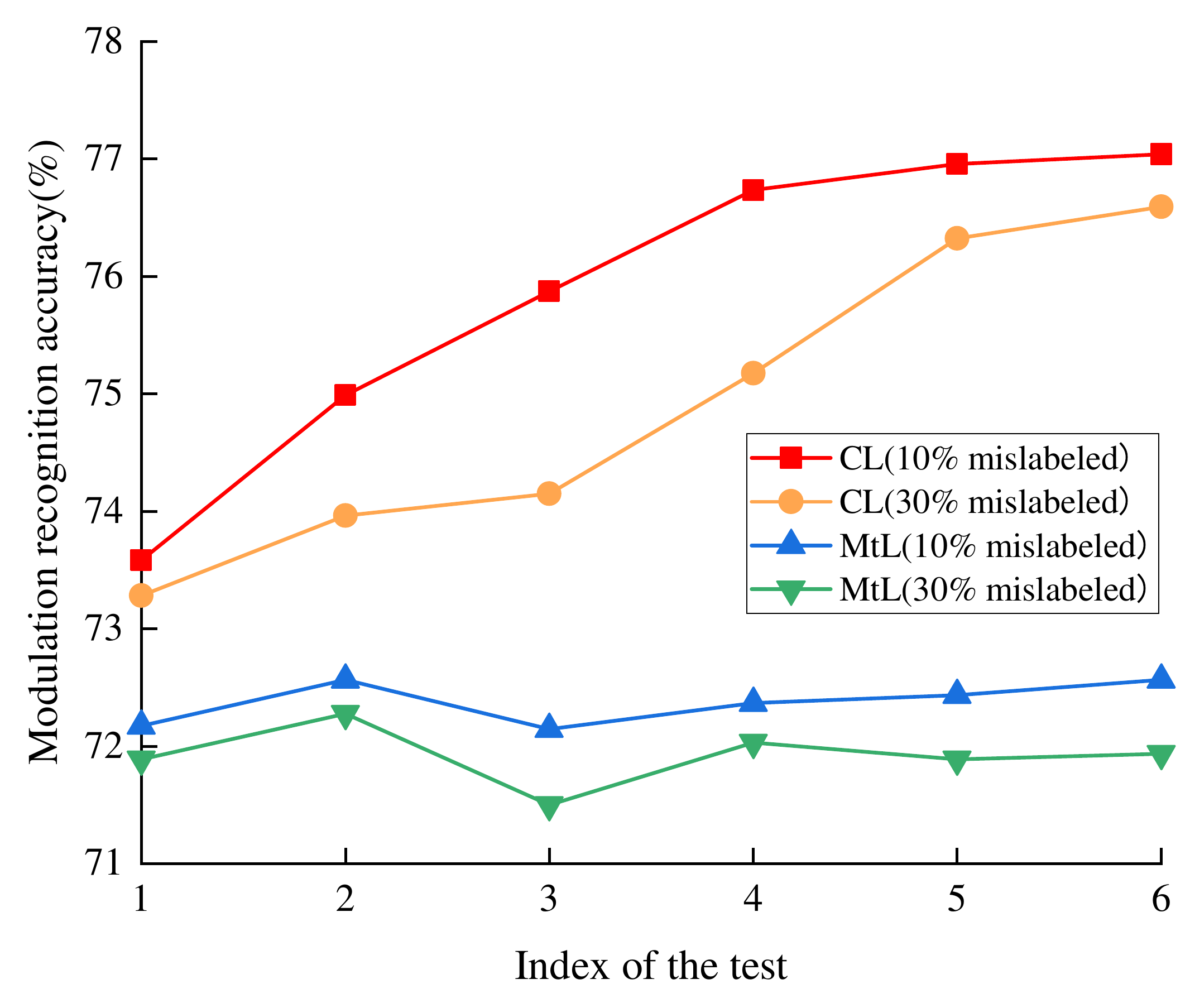}
        \label{Fig.badsamples.d}
	}
\caption{{\textbf{Performance of the algorithm selection and hyper-parameter selection with mislabeled training samples.} \textbf{a}, Algorithm selection accuracy. \textbf{c}, Hyper-parameter selection accuracy. \textbf{b} and \textbf{d}, modulation recognition accuracy achieved by the selected algorithm and hyper-parameters respectively.}
\label{Fig.badsamples}}
\end{figure}

  In this case, a better algorithm type or hyper-parameters can be selected and a better matching relationship between features of the environment and tasks and the selection of an appropriate  algorithm type or hyper-parameters can be established and stored in the cognitive case space, which are helpful to improve the training results. On the other hand, the ratio of the mislabeled cases in the cognitive case space decreases with the index of the test since more better matching relationship between features of the environment and tasks and the selection of an appropriate  algorithm type or hyper-parameters are stored in the cognitive case space when the test increases. Those results demonstrate that our proposed CL framework has the capability of ``good money driving out bad money''.  However, for the MtL framework, the ratio of the mislabeled cases is a constant during the learning process and thus the performance cannot be improved with the increase of the index of the test.}
\section{Discussion and Conclusion}
A CL framework was proposed for the dynamic wireless environment and tasks motivated by the brain cognitive mechanism. Our proposed CL framework also has two processes, namely, the online process and offline self-learning process. The online process quickly performs the selected algorithm on the data and responses to the environment while the offline self-learning process improves the performance by selecting more appropriate algorithm and hyper-parameters. The mathematical framework was established for our proposed CL framework. Simulation results have demonstrated that our proposed CL framework has three advantages, namely, the capability of adapting to the dynamic environment and tasks, the self-learning capability, and the capability of ``good money driving out bad money''. Our proposed CL framework enriches the machine learning frameworks and open many new avenues for future machine learning investigations.

{\color{blue}Our future works will focus on two aspects. On one hand, we will study how to exploit our proposed framework to tackle the issues existing in the wireless communications, such as resource allocation, interference classification, etc. On the other hand, we will investigate the data and knowledge dual-driven learning  mechanism facilitated by our proposed  framework since the cognitive case space contains knowledge.}

\end{document}